# Transfer Learning as an Enabler of the Intelligent Digital Twin

Benjamin Maschler [a,b,*], Dominik Braun [a,b], Nasser Jazdi [a,b], Michael Weyrich [a]

[a] *University of Stuttgart, Institute of Industrial Automation and Software Engineering, Pfaffenwaldring 47, 70569 Stuttgart, Germany*
[b] *These authors contributed equally to this publication.*

* Corresponding author. Tel.: +49 711 685 67295; Fax: +49 711 685 67302. E-mail address: benjamin.maschler@ias.uni-stuttgart.de

**Abstract**

Digital Twins have been described as beneficial in many areas, such as virtual commissioning, fault prediction or reconfiguration planning. Equipping Digital Twins with artificial intelligence functionalities can greatly expand those beneficial applications or open up altogether new areas of application, among them cross-phase industrial transfer learning. In the context of machine learning, transfer learning represents a set of approaches that enhance learning new tasks based upon previously acquired knowledge. Here, knowledge is transferred from one lifecycle phase to another in order to reduce the amount of data or time needed to train a machine learning algorithm.

Looking at common challenges in developing and deploying industrial machinery with deep learning functionalities, embracing this concept would offer several advantages: Using an intelligent Digital Twin, learning algorithms can be designed, configured and tested in the design phase before the physical system exists and real data can be collected. Once real data becomes available, the algorithms must merely be fine-tuned, significantly speeding up commissioning and reducing the probability of costly modifications. Furthermore, using the Digital Twin's simulation capabilities virtually injecting rare faults in order to train an algorithm's response or using reinforcement learning, e.g. to teach a robot, become practically feasible.

This article presents several cross-phase industrial transfer learning use cases utilizing intelligent Digital Twins. A real cyber physical production system consisting of an automated welding machine and an automated guided vehicle equipped with a robot arm is used to illustrate the respective benefits.

## 1. Introduction

In recent years, the practical application of artificial intelligence algorithms in networked production systems has been the focus of numerous scientific publications [1]. Many of these aim at providing new functionalities as an addition to conventional control software, e.g. in order to detect anomalies [2], optimize operating parameters [3] or predict faults [4]. However, integrating those into live industrial systems remains a challenge, because the required training datasets are hard to acquire [5, 6].

One possible vehicle to convey these new concepts into actual industrial applications would be the intelligent Digital Twin [7]. Together with lifecycle-spanning, so-called cross-phase transfer learning [5], AI algorithms could be designed and trained in simulations and then swiftly deployed onto the real assets without requiring as much training data or time.

*Objective*: In this article, case studies from the domain of industrial manufacturing underlining the potentials of intelligent Digital Twins enhanced with cross-phase transfer learning capabilities are presented and analyzed.

*Structure*: In chapter 2, related work on the topics of intelligent Digital Twins and transfer learning is introduced. From there, a concept is derived in chapter 3. Chapter 4, then, presents case study scenarios as well as the cyber-physical production system and its Digital Twin used therein. Finally, a conclusion and an outlook are given in chapter 5.

## 2. Related Work

*2.1. Intelligent digital twin*

The Digital Twin is a well-established concept which is defined differently by each author depending on the use case and the discipline. Nevertheless, there are similarities in certain core elements which are the same regardless of the use case and which define the Digital Twin in general. They have been identified by literature survey presented in [7]:

The most fundamental part of any Digital Twin are the asset's digital models, e.g. CAD, ECAD, simulation models, software models and many more. Which of those a specific Digital Twin requires strongly depends on the real asset as well as the use case and can therefore differ between Digital Twins and over their lifecycle [8, 9]. Furthermore, the relations between different models (instance-instance relation), within a structure (parent child relation) and the dependencies of elements (inheritance relations) are of great importance [10]. Together, these models and their relations form the *digital replica* (see Fig 1).

Besides these elements, a digital replica needs to have a set of three additional abilities to form a Digital Twin: an active data acquisition from the real manufacturing system and its environment to replicate the dynamic behavior and to adapt to the current status of the system; an interface to synchronize the models and their relations with any modifications occurring to the real manufacturing system; an interface for co-simulation with other Digital Twins to represent a complex factory commonly constituted by more than a single Digital Twin [7]. Such a set of models and abilities enriched by organizational information and meta data is then called *Digital Twin* (DT).

There are many applications for a DT of manufacturing systems, in the areas of planning, control or maintenance, like virtual commissioning or fault tracing [11]. In general, the advantage of having a Digital Twin in these use cases as well as all others is that the real plant, the physical asset, is not needed [12]. All tests, optimizations, programming and maintenance can be done virtually and supported by computing power. This shortens downtime or the time to ramp-up. Thus, the system can be used continuously and operation does not have to be interrupted for planning or testing. The DT can be used for support during the hole lifecycle of a plant starting from its design over engineering, operation up to optimization as long as the Digital Twin is continuously adapted to the real asset [13].

The *intelligent Digital Twin* is an extension of the DT and enriches it with artificial intelligence to automatically adapt models, provide benefits and new abilities for the physical system and its environment. By definition, the intelligent DT is capable to observe and analyze the environment which is necessary to learn and gain new information from it [14].

Examples for these new abilities are fault prediction, anomaly detection, control code generation, flexible control, machine condition evaluation or production sequence optimization [7, 11]. The intelligent DT observes the environment using the operating data (sensor and actuator values) and analyzes it together with the models of the DT to gain new knowledge beyond the explicitly defined information in the existing models. In the case of an intelligent DT, the artificial intelligence algorithms are an integral part and must have an information coupling to the models and process data of the DT.

So far, there are only standalone implementations which extract and analyze information from the digital twin without any feedback to or re-integration into the DT. This might be due to perceived technical obstacles in transferring knowledge back into the DT and updating it again whenever the need arises.

*2.2. Transfer learning*

In the field of machine learning, transfer learning refers to the transfer of knowledge and skills from previously learned tasks to new tasks in order to improve performance on the latter [15]. Although a lot of literature on this topic focusses on the application of transfer learning on visual recognition (esp. in the medical domain), natural language processing or communication tasks [16, 17], there has recently been increasing interest from the industrial domain as well [5, 18].

Here, transfer learning is meant to solve two problems hindering a more widespread deployment of deep learning techniques:

- Due to only very low numbers of identical industrial machinery, high standards of data protection and low levels of cooperation between different enterprises, sufficiently large datasets and diverse for successful training are hard to acquire [19].
- Due to increasing demand for frequent reconfigurations [20], changing processes and dynamic environments, quickly outdating datasets once acquired only provide short-term representations of the problem space necessitating continuous data collection and algorithm retraining [21].

Transfer learning offers mitigation to those challenges by enabling algorithms to train not only on datasets characterizing the task at hand, but on related ones, e.g. from other lifecycle phases, as well. Furthermore, it allows algorithms to adapt to changing tasks without requiring retraining from scratch.

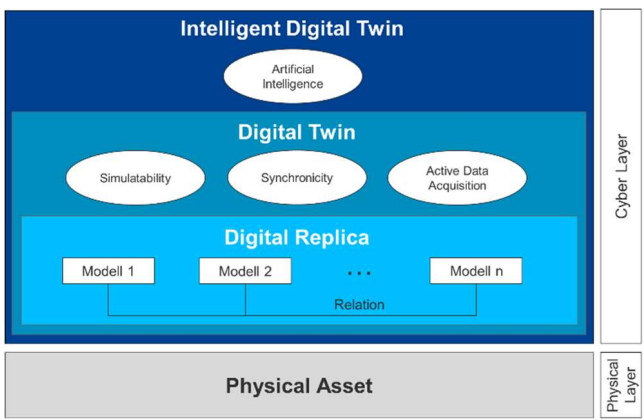

Fig. 1. Schematics of the relations between the intelligent digital twin, digital twin and digital replica according to [7]

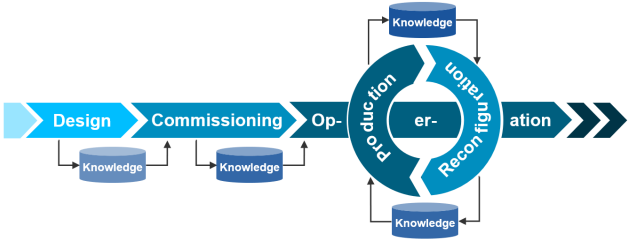

Fig. 2. Cross-phase industrial transfer learning applied to an excerpt of the industry 4.0 lifecycle according to [22]

To give some examples, in [23], the authors created an algorithm capable of single-shot object recognition learning on dynamic tasksets. In [24], an anomaly detection algorithm tolerating changing input data dimensionality is presented. In [6], the authors propose to pre-train a quality management algorithm on simulated data before fine-tuning it on the actual dataset collected from the real process.

This transfer of knowledge between different modes of data, i.e. simulated and real data, or rather different lifecycle phases used in the last example is termed 'cross-phase industrial transfer learning'. It represents one of four base-use cases of industrial transfer learning according to [5] (see Fig. 2). So far, only very few publications address this use case. This might be due to a perceived lack of beneficial applications or a lack of high-quality datasets combining data collected from simulated and real assets.

## 3. Concept

Models are a means to replicate parts of a real asset and store this information, e.g. a DT's models contain information describing the physical asset and its environment. Algorithms from the field of artificial intelligence could utilize this information in order to deduct knowledge regarding the physical asset's behavior and its interaction with the environment.

More specifically, executing these models by simulation could make this information accessible to deep neural networks by providing them with training data. Obviously, this would only be beneficial if those algorithms can be trained on simulation data in advance and then their knowledge be transferred to the real asset without much time needed.

On the other hand, transfer learning directly used on algorithms that are deployed on real assets might pose safety or security issues as the exact results of such a transfer are not predictable. Thus, even if a real asset could provide a direct opportunity for knowledge transfer, it might be preferable to take a de-tour into the virtual world in order to test the resulting system in simulation first.

We therefore propose to combine the two concepts of intelligent DT and transfer learning in order to greatly advance their applicability as well as functionality.

## 4. Case study

In this chapter, we present case studies in which the potential of the aforementioned intelligent DT capable of cross-phase transfer learning is analyzed. To this purpose, we introduce an actual production system and its digital twin as an example.

### 4.1. Cyber-physical production system

A cyber-physical production system is used to investigate and demonstrate the possibilities of flexible production systems and their DT. It was designed and built using actual industrial machinery on the research campus ARENA2036 (Active Research Environment for the Next Generation of Automobile). It consists of three automated units (welding machine, movable robot and an intelligent warehouse) which each have their own decentralized control unit to control and provide the respective services. An additional head control unit is responsible for coordinating the units to produce a model car from four sheet metal parts (see Fig. 3).

The following functionalities are implemented: The intelligent warehouse provides prefabricated metal parts in workpiece carriers and withdraws empty carriers. The material flow inside this warehouse is controlled by 24 actuators and monitored by 37 sensors, which represent the current state of occupation. This information is made available to the other system participants by the warehouse PLC via industrial WLAN. The movable robot collects the metal parts in carriers from the warehouse, assembles the model car and transfers it to the welding machine. Because of this connection by mobile robot which acts as a driverless transport vehicle, the units do not have to be arranged in a fixed, linear chain. Furthermore, not only the robot but also the warehouse is equipped with wheels and can therefore easily be moved which enables a variable and easily modifiable production process.

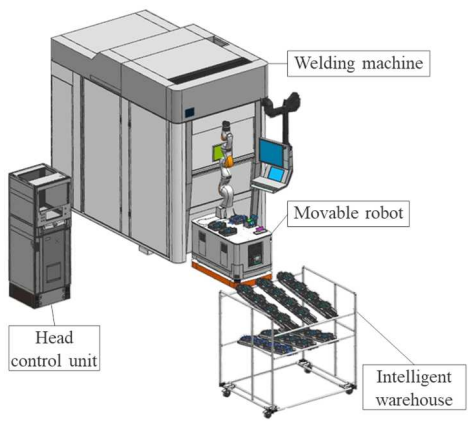 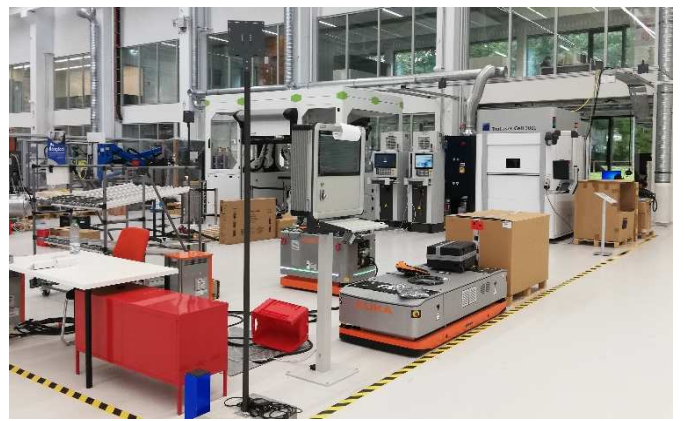

Fig. 3. Cyber-physical production system at the ARENA2036 (left: CAD model; right: photo of real asset)

The cross-domain models of the DT were created during the design phase using a PLM system and is the cyber part of the production system. These PLM systems are able to integrate and manage numerous tools and their models during design, operation and maintenance. The DT of the system consists of mechanical models, kinematic definitions, electrical models, process simulations and automation software models. It was used to, e.g., virtually verify in advance that the robot could reach the carriers and the clamping fixtures inside the welding machine.

*4.2. Scenario 1: Design of deep neural networks*

In the design of deep learning algorithms based upon deep neural networks, many parameters (e.g. types and numbers of nodes, topology, data preprocessing) need to be fitted to the specific use case at hand. Studies show that the quality of this fit greatly influences the resulting algorithm's performance [25]. Furthermore, the factors that support or hinder the adaption of algorithms proven to solve similar use cases are still largely unknown [21], making diligent testing of proposed solutions a key priority.

If such testing and the ensuing redesign of a deep learning algorithm could be carried out prior to the assembling of the actual system, i.e. the real asset, one would save valuable time in the commissioning phase. This can be achieved by the proposed enhancement of intelligent DT with transfer learning:

Through the use of models incorporated in the DT and data generated by executing them, one can make and test the necessary design choices regarding the deep neural network virtually during design phase (see Fig. 4). Naturally, the transferability of the resulting algorithm then depends on whether the DT does indeed resemble the real asset in all important aspects.

**Example**: The DT of our cyber physical production system (see chapter 4.1) shall be equipped with an anomaly detection algorithm, e.g. based upon LSTM-networks. In order to decide on the relevant parameters (e.g. number of nodes, number of layers, learning rate), tests are carried out using the DT already available at late stages in the design phase. Transfer learning is then used to safeguard the algorithm's adaption from simulation to real asset.

Because all basic features and functionalities of the real asset are included at this stage, the resulting algorithm will – with a high degree of certainty – be able to perform on the real asset as well. Even today, such assumptions are commonly made, e.g. in the case of virtual commissioning, which also depends on a high degree of similarity between DT and real asset.

*4.3. Scenario 2: Deployment of deep neural networks*

In addition to – but not necessarily building on - Scenario one (see chapter 4.2), valuable time can also be saved in the deployment of deep neural networks:

When deep learning algorithms are trained, large and diverse datasets are needed but hard to acquire (see chapter 2.2). Even after commissioning, generating training datasets might require months-long data collection runs in which the algorithm would still be untrained and therefore functionally unavailable.

If pre-training could be carried out prior to commissioning, one would reduce the amount of data needed from the real asset, thereby shorten the duration of data collection runs and speeding up the availability of the fully trained algorithm. This can be achieved by the proposed enhancement of intelligent DT with transfer learning:

Through the use of models incorporated in the DT in simulations to generate a training dataset, one can pre-train the deep neural network virtually during design phase. Once the real asset is assembled and some data collected, this algorithm would merely need to be fine-tuned, which requires far less data than the complete training would have (see Fig. 5).

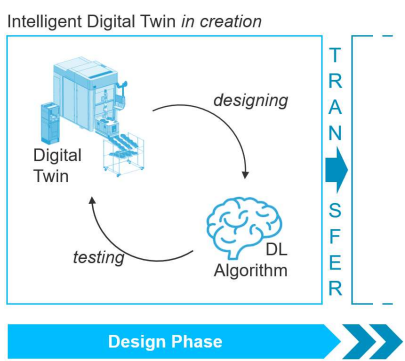

Fig. 4. Design of deep learning (DL) algorithms for the intelligent digital twin during design phase (see Scenario 1)

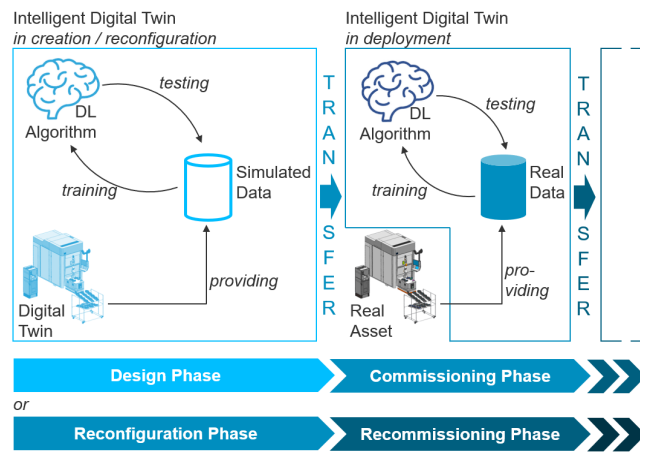

Fig. 5. Training of deep learning (DL) algorithms for the intelligent digital twin during design and commissioning or reconfiguration and recommissioning phase (see Scenario 2)

This feature can be used not only in the initial design and commissioning, but also in the event of reconfiguration and re-commissioning – always allowing a virtual pre-training to speed up the roll-out of the final, fully functional algorithm.

**Example**: The DT of our cyber-physical production system (see chapter 4.1) shall still be equipped with an anomaly detection algorithm. In order to pre-train it, simulations are carried out using the DT already available at late stages in the design phase. During these simulations, data describing the desired system behavior is collected and the algorithm trained with it. As soon as the real asset is available, a smaller training dataset is collected from it in matter of a few hours, facilitating the algorithm's adaption to the real asset using fine-tuning based on transfer learning. Furthermore, this allows for the automatic adaption of models to changes possibly occurring during commissioning.

*4.4. Scenario 3: Injection of rare faults*

Further extending Scenario two (see chapter 4.3) the overall performance and robustness of the algorithm ca be increased additionally to the time savings.

One challenge in training data collection is achieving the necessary diversity (see chapter 2.2): In order to train an algorithm's behavior for events involving (rare) faults, one needs data describing such faults. Using the real asset, this data might not be obtainable as the faults might occur to seldomly or might be too much of a safety or security risk for them to be allowed to occur.

The proposed enhancement of intelligent DT with transfer learning solves this problem: Using the models incorporated in the DT in simulations to generate a training dataset, one can inject those faults into the virtual system in order to generate corresponding training data [26] (see Fig. 6). Again, the transferability of the resulting, trained algorithm depends on whether the DT does indeed resemble the real asset in all important aspects – an assumption readily made in other contexts. A remaining challenge is the evaluation of results acquired this way.

**Example**: We want to make sure that the intelligent DT of our cyber-physical production system's (see chapter 4.1) anomaly detection algorithm reliably detects some specific faults, e.g. rare sensor failures or dangerous sabotage by coating the metal parts with a combustible substance. We therefore inject those faults into the simulation running the DT's models and collect the resulting data to test – and potentially retrain – the algorithm. Such an algorithm can then be adapted to the real asset using transfer learning in order to prevent the error, e.g. by conducting counter measures, or to work towards tolerating the error. Furthermore, we could try to create more (rare) situations addressing the algorithm's known weaknesses.

*4.5. Scenario 4: Enabling reinforcement learning*

Whereas Scenarios one to three are today only seldomly described in literature, studies regarding reinforcement learning in simulated environments are widely available. However, they usually do not highlight their implicit reliance on the concepts of DT and transfer learning:

When reinforcement learning is used, acquiring a fully trained deep learning algorithm can take a long time as the algorithm finds the optimal parameters by trial and error, merely guided by feedback information. Even on simple tasks this can take thousands of trials, costing considerable amounts of time [27]. Furthermore, if conducted on the real asset, all unsuccessful trials occur in reality, leading to hundreds of unwanted system states, potentially even harmful ones.

Transferring reinforcement learning into the virtual realm is therefore necessary to enable its wide-spread utilization. Naturally, such a virtual realm needs to feature the necessary models and be simulable. Furthermore, in order to be able to transfer the results back onto the real asset, this virtual realm needs to be able to synchronize itself with the real asset and collect data on its behavior – thus, it needs to be a DT. Featuring artificial intelligence functionalities, it needs to be an intelligent DT and to bridge the gap between the digital models and the real asset, it needs transfer learning capabilities (see Fig. 7).

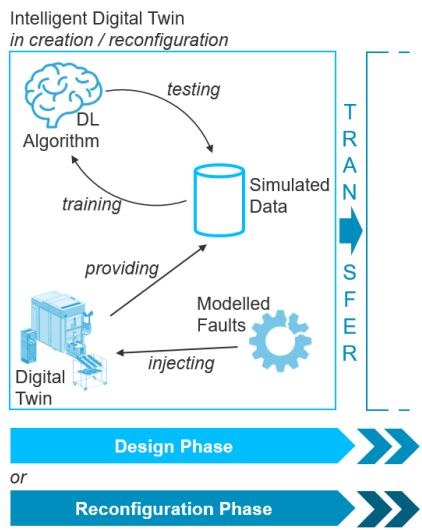

Fig. 6. Injection of rare faults into the intelligent digital twin during design or reconfiguration phase (see Scenario 3)

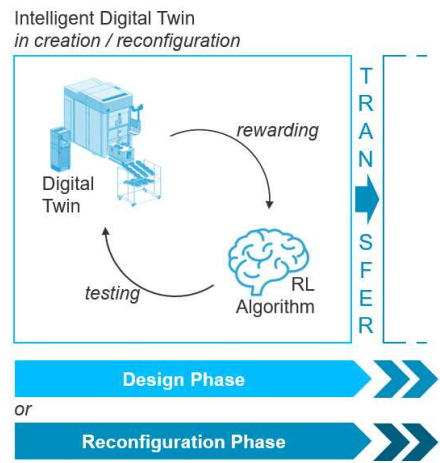

Fig. 7. Reinforcement learning (RL) conducted inside the intelligent digital twin during design or reconfiguration phase (see Scenario 4)

**Example**: There are numerous examples on this topic. For our cyber-physical production system (see chapter 4.1), training the robot arm for picking up the metal parts and assembling the model car can be done using reinforcement learning on the digital twin. Transfer learning is then used to bridge the gap between simulation and real asset. This speeds up the process, e.g. by learning on multiple instances parallelly, and avoiding errors in real life, e.g. dropped metal parts that would have to be picked up by a human operator.

## 5. Conclusion

The presented scenarios highlight potential benefits that arise from enhancing intelligent digital twins with cross-phase transfer learning when deep learning algorithms are used: Transfer learning allows deep learning algorithms to be designed, trained and possibly enhanced with fault injection using the digital twin's simulation capabilities and then to be adapted onto the real asset. This leads to time savings or performance increases. Evaluations of the specific extend of these benefits are still ongoing and will be included in later publications.

Furthermore, it could be demonstrated that the necessary, underlying assumption of similarity between simulation and reality can readily be made if the digital twin used is detailed enough. In fact, even today it already is frequently made whenever reinforcement learning algorithms are used.

Concludingly, intelligent digital twins would greatly profit from a more wide-spread utilization of the proposed concept – in terms of expanded functionality, reduced training efforts and increased robustness.

Although the very first implementations outside of the field of reinforcement learning have been made, there still is a great need for further research regarding a greater variety of reference implementations, benchmark comparisons and more real-life applications.